\documentclass[conference]{IEEEtran}
\IEEEoverridecommandlockouts
\usepackage{cite}
\usepackage{amsmath,amssymb,amsfonts}
\usepackage{algorithmic}
\usepackage{graphicx}
\usepackage{textcomp}
\usepackage{xcolor}
\def\BibTeX{{\rm B\kern-.05em{\sc i\kern-.025em b}\kern-.08em
    T\kern-.1667em\lower.7ex\hbox{E}\kern-.125emX}}

\usepackage{booktabs,multirow}
\definecolor{gray}{rgb}{0.5, 0.5, 0.5}
\usepackage{arydshln} 

\usepackage{xurl} 
\usepackage{url}
\usepackage[hidelinks]{hyperref}

\usepackage[caption=false,font=footnotesize]{subfig}

\makeatletter

\begin{document}



\title{TGIF: Text-Guided Inpainting Forgery Dataset\\
\thanks{ 
This work was funded by the Flemish government’s Department of Culture, Youth \& Media (under the COM-PRESS project), by IDLab (Ghent University -- imec), by Flanders Innovation \& Entrepreneurship (VLAIO), by Research Foundation -- Flanders (FWO) (V419524N), and by the European Union under the Horizon Europe vera.ai project (grant number 101070093).
}
}

\author{
\IEEEauthorblockN{Hannes Mareen\IEEEauthorrefmark{1}, Dimitrios Karageorgiou\IEEEauthorrefmark{2}, Glenn Van Wallendael\IEEEauthorrefmark{1}, Peter Lambert\IEEEauthorrefmark{1}, Symeon Papadopoulos\IEEEauthorrefmark{2}
}
\IEEEauthorblockA{
\IEEEauthorrefmark{1} \textit{IDLab, Ghent University -- imec, Ghent, Belgium, e-mail: firstname.lastname@ugent.be} \\
\IEEEauthorrefmark{2} \textit{Information Technologies Institute, CERTH, Thessaloniki, Greece, e-mail: dkarageo@iti.gr, papadop@iti.gr} 
}
}

\maketitle

\begin{abstract}
Digital image manipulation has become increasingly accessible and realistic with the advent of generative AI technologies. Recent developments allow for text-guided inpainting, making sophisticated image edits possible with minimal effort.
%
This poses new challenges for digital media forensics. For example, diffusion model-based approaches could either splice the inpainted region into the original image, or regenerate the entire image. 
In the latter case, traditional image forgery localization (IFL) methods typically fail.
%
This paper introduces the Text-Guided Inpainting Forgery (TGIF) dataset, a comprehensive collection of images designed to support the training and evaluation of image forgery localization and synthetic image detection (SID) methods. The TGIF dataset includes approximately 75k forged images, originating from popular open-source and commercial methods, namely SD2, SDXL, and Adobe Firefly.
We benchmark several state-of-the-art IFL and SID methods on TGIF. 
%
Whereas traditional IFL methods can detect spliced images, they fail to detect regenerated inpainted images.
Moreover, traditional SID may detect the regenerated inpainted images to be fake, but cannot localize the inpainted area.
Finally, both IFL and SID methods fail when exposed to stronger compression, while they are less robust to modern compression algorithms, such as WEBP.
%
In conclusion, this work demonstrates the inefficiency of state-of-the-art detectors on local manipulations performed by modern generative approaches, and aspires to help with the development of more capable IFL and SID methods.
The dataset and code can be downloaded at \url{https://github.com/IDLabMedia/tgif-dataset}.
\end{abstract}


\section{Introduction}
\label{sec:intro}
Image manipulation is becoming more accessible with recent image editing tools based on generative AI (GenAI) achieving outstanding levels of photorealism~\cite{zhang2023text, zhan2023multimodal}. The advent of open-source and commercial text-guided approaches, such as Stable Diffusion (SD)~\cite{rombach2022sd} and Adobe Firefly, introduced the ability to locally manipulate an image using simple text prompts, even to non-experts. However, as high-quality image manipulation is becoming more accessible, there are great risks to be exploited for malicious purposes such as disinformation, fraud, and fake evidence. Therefore, image forensic methods aim to detect and localize manipulations~\cite{mehrjardi2023survey, tariang2024synthetic}.
While inpainting has long been part of the target manipulation types of state-of-the-art image forgery localization (IFL) methods, it is now revived due to the proliferation of text-to-image generative diffusion models~\cite{zhan2023multimodal}. Previously, inpainting meant the removal of objects~\cite{guan2019mfc,mahfoudi2019defacto}, whereas text-guided inpainting models enable the addition or adaptation of objects.

Text-guided inpainting using diffusion-based GenAI models works differently than traditional inpainting: 
an image (crop) is given as input to a model, along with a mask of the inpainted area and a textual prompt. Through a series of diffusion and denoising steps, the model regenerates the entire image, yet only visibly changing the inpainted area~\cite{rombach2022sd}.
There are two ways of using text-guided inpainting. First, by splicing the inpainted area into the original image, which is most useful when the generative model does not support high resolutions.
Second, the fully regenerated image can be used as is.
With new models supporting larger resolutions, the latter case is becoming more practical.
However, localizing the inpainted region when most forensic cues have been corrupted due to the regeneration of the whole image is a new and hard challenge for IFL algorithms, and untouched by synthetic image detection (SID) methods.

Training and evaluating forensic methods requires large and varied datasets. Although several image manipulation datasets exist, text-guided inpainting is typically not included in them.
In fact, to the best of our knowledge, the only dataset including text-guided inpainting is COCO-Glide~\cite{guillaro2023trufor}, consisting of 512 manipulated images of $256\times 256$ px, created using a single method, i.e. GLIDE~\cite{nichol2021glide}. The dataset lacks extensive metadata, and does not vary in terms of splicing type and mask type.
Hence, there is a need for larger text-guided inpainting datasets that can be utilized for both training and evaluation, consisting of high-resolution images.

This paper presents the Text-Guided Inpainting Forgery (TGIF) dataset.
The contribution of this work is two-fold.
First, we introduce our TGIF dataset in Section~\ref{sec:dataset}, which can be used for both training and evaluation. The dataset consists of approx. 75k fake images, with resolutions up to $1024\times 1024$ px that we generated using three inpainting models (SD2, SDXL \& Adobe Firefly).
Additionally, we provide both the manipulated image where the inpainted area is spliced in the original image, as well as the fully-regenerated image, when possible.
We provide metadata about the estimated aesthetic quality, as well as image-text matching scores that indicate whether the generated image aligns with the given prompt.

Second, we evaluate our TGIF test set on state-of-the-art IFL and SID methods, in Section~\ref{sec:benchmark}. Additionally, we study the effects of JPEG and WEBP compression on detection performance. In this way, we provide valuable insights into the text-guided inpainting manipulations, how to detect them in practice, and highlight open challenges.

\section{Related Work}
\label{sec:related}


\subsection{Image Forgery Localization (IFL)}
\label{sec:related-methods-ifl}
To detect local image manipulations, in which a region of an image was forged whereas the rest remains pristine, a variety of methods have been proposed~\cite{mehrjardi2023survey}. In general, IFL methods either aim to detect imperceptible inconsistencies in the image, such as the artifacts originating from compression~\cite{mareen2022comprint} or the noise pattern introduced by camera sensors~\cite{Cozzolino2019Noiseprint}, or they aim to detect visible ones, such as boundaries between the forged and pristine regions~\cite{dong2022mvss}, while others fuse several forensic cues for increased robustness~\cite{triaridis2024mmfusion, karageorgiou2024fusion, mareen2023harmonizing}. Furthermore, there exist a few IFL methods that specifically target inpainting~\cite{li2017localization, wu2021iid, ijcai2021noisedoesntlie}, which exploit specific changes made by inpainting models, such as in image Laplacian~\cite{li2017localization}, or which automatically learn to extract relevant features~\cite{wu2021iid, ijcai2021noisedoesntlie}.
However, they have mostly been superseded by more recent general works~\cite{liu2022pscc, guillaro2023trufor}.

Recent well-performing methods~\cite{liu2022pscc, hu2020span, wu2022ImageForensicsOSN, dong2022mvss, wu2019mantranet, kwon2022catnetv2, guillaro2023trufor} are mostly based on deep learning, i.e., they train and evaluate on datasets containing diverse manipulation types, such as splicing, copy-move, and ``traditional'' inpainting/removal, in order to detect a broad range of alterations. 
However, such methods have not been trained to detect fully regenerated images using text-guided generative models. 

\subsection{Synthetic Image Detection (SID)}
\label{sec:related-methods-sid}
Early popular synthetic image generation approaches were mostly based on Generative Adversarial Networks (GANs), whereas newer ones typically rely on diffusion models (DMs).
Accordingly, several detection methods were trained on images generated by GANs~\cite{wang2019cnngenerated, frank2020fredetect, ojha2023univFD, liu2020gramnet, tan2023LGrad, tan2024NPR, zhong2023patchcraft}, whereas newer ones typically focus on DMs or a mix of both~\cite{corvi2023dimd, wang2023dire, koutlis2024Rine, sha2023defake}.

In general, most methods aim to extract features or a fingerprint, and somehow compare them to those typically observed in synthetic images.
The type of feature may have a large influence on the model's generalization performance. For example, some methods were trained on GANs, yet demonstrated the ability to detect images generated by DMs~\cite{ojha2023univFD, tan2024NPR, zhong2023patchcraft}.
However, state-of-the-art SID methods were trained and evaluated on real and synthetic images, lacking support for images that were inpainted and hence contain both fake and real parts (albeit regenerated by DMs). In other words, they lack the ability of localizing regions inpainted by DMs. 

\subsection{Image Forensics Datasets}
\label{sec:related-datasets}
A variety of datasets for IFL and SID exist, and have been listed in recent overview papers~\cite{verdoliva2020media, lin2024detecting}.
%
Most relevant to this paper, inpainted images can be found in the NIST16 \cite{guan2019mfc} and DEFACTO \cite{mahfoudi2019defacto} datasets. However, the inpainting in them is not guided by a textual prompt, but instead aims to simply remove the selected object through splicing.

To the best of our knowledge, the only dataset that includes text-guided inpainting is COCO-Glide~\cite{guillaro2023trufor}, which consists of 512 manipulated images with a resolution of $256 \times 256$ px, cropped from the MS-COCO dataset and inpainted using GLIDE~\cite{nichol2021glide}.
This dataset is small and mainly useful for evaluation. Additionally, the relatively low resolution is already outdated, with new models supporting resolutions up to $1024 \times 1024$ px.
Moreover, it neither considers the differences between splicing the generated content into the inpainted region of the original image and re-generating the entire picture with the GenAI model, nor does it assess the perceptual quality of the inpainted images. 
Thus, the forensics community lacks a general large dataset that considers local image manipulations performed by generative models, while including localization masks for images that have been fully regenerated.

\section{TGIF Dataset}
\label{sec:dataset}
This section presents our text-guided inpainting forgery dataset, containing both spliced and fully regenerated images; yet only visually different in the inpainted area.

\subsection{Source of Authentic Data}
\label{sec:dataset-generation-source}
We used the MS-COCO dataset~\cite{lin2014coco} (\emph{val2017}) to collect authentic images licensed under a Creative Commons Attribution 4.0 License, along with captions and object masks (segmentation mask and bounding box). The images contain objects from 80 categories (such as types of animals, food, and furniture). Of those, we excluded \emph{knife} and \emph{frisbee} because these prompts were not allowed by Adobe Firefly.


We edited the provided Flickr URLs in order to collect images with dimensionality of up to 1024 pixels (the largest publicly available resolution)~\cite{flickr_services_photo_image_urls}, as the original COCO metadata linked to Flickr images with a size limited to 640 pixels on their largest dimension.

From each of the object classes, we selected a maximum of 50 images.
We excluded images in cases where the 1024p Flickr URL was not accessible, as well as those with one of their dimensions smaller than 512 pixels. Moreover, the object bounding box resolution should be smaller than 512$\times$512 px, and the product of width and height should be larger than $64^2$.

\subsection{Text-Guided Inpainting Generation}
\label{sec:dataset-generation-inpainting}
For each authentic image and depicted object, we generated inpainted variants, where the authentic object was replaced by a generated one of the same class. For example, a real cat was replaced by a generated cat.

We utilized three text-guided inpainting methods: a) the open-source SD2 (\emph{stabilityai/stable-diffusion-2-inpainting}), which constitutes a first-generation high-fidelity diffusion model, b) the open-source SDXL (\emph{diffusers/stable-diffusion-xl-1.0-inpainting-0.1}), representing the second generation and supporting images of higher resolution, and
c) the commercial Adobe Firefly (in Adobe Photoshop 25.4.0). The open-source models are encountered in popular inpainting pipelines, such as Inpaint-Anything~\cite{yu2023inpaintanything}, while the commercial approaches are more accessible to non-experts.
These three generative approaches constitute a representative set of the available approaches for local manipulation through generative models.

For SD2 and SDXL, we cropped images (centered around the object of interest) to 512$\times$512 px and 1024p, respectively, as these are their respective native resolutions.
Additionally, the dimensions needed to be a multiple of 8.
In Adobe Firefly, we did not need to take any crops, although this may happen internally.
To perform the inpainting, a mask of the object is given to select the area to be inpainted. For this, we utilized the segmentation mask and bounding box provided by COCO.
Moreover, each inpainting operation of each generative model generated was repeated three times in a batch, resulting in three output variations (set with the \emph{num\_images\_per\_prompt} variable, which internally uses a different seed for each variation), which we all included in the dataset. I.e., each image was inpainted using two masks, and each of those inpaintings generated three variations. This process was repeated for each of the three inpainting methods.

As a textual prompt in SD2 and SDXL, we utilized the object category followed by a caption from the image (i.e., describing the full image). During initial experiments, we noticed that including the caption resulted in better results, which was also observed in related work~\cite{xie2023smartbrush, wang2023imagen}.
However, text misalignment between the prompt and the inpainted area could still occur.
For Adobe Firefly, we only included the object category as prompt (i.e., only describing the inpainted area), as we noticed that this was sufficient in our initial experiments.

We randomly chose several other parameters. For each image, the number of inference steps is a random integer between 10 and 50, and the guidance scale is a random float between 1 and 10. The seed is randomly chosen. These parameters are saved as metadata.

When inpainting using GenAI, the entire input image is regenerated. Although only the masked area has visible changes, also the non-masked area is (imperceptibly) changed by the GenAI algorithm, since the entire image undergoes diffusion.
In Adobe Firefly (Photoshop), the GenAI inpainted image is spliced back into the original image, using the provided mask with an additional small gradient border (i.e., a few pixels larger).
This is a similar approach as provided by other inpainting pipelines, such as the one provided in HuggingFace diffusers~\cite{huggingface_diffusers}.
For consistency, we spliced the SD2 inpainted image into the original image using the same adapted mask from Firefly, with an additional gradient border.
Additionally, we saved the fully regenerated output image as a 512$\times$512 px image.
For SDXL, we only saved the fully regenerated output image as a 1024p image. That is because the SDXL inpainting model discolors the entire image, which is extremely visible when splicing into the original (a known issue~\cite{diffusers_github_issue}).

\begin{figure*}
\centering
\includegraphics[width=0.86\linewidth]{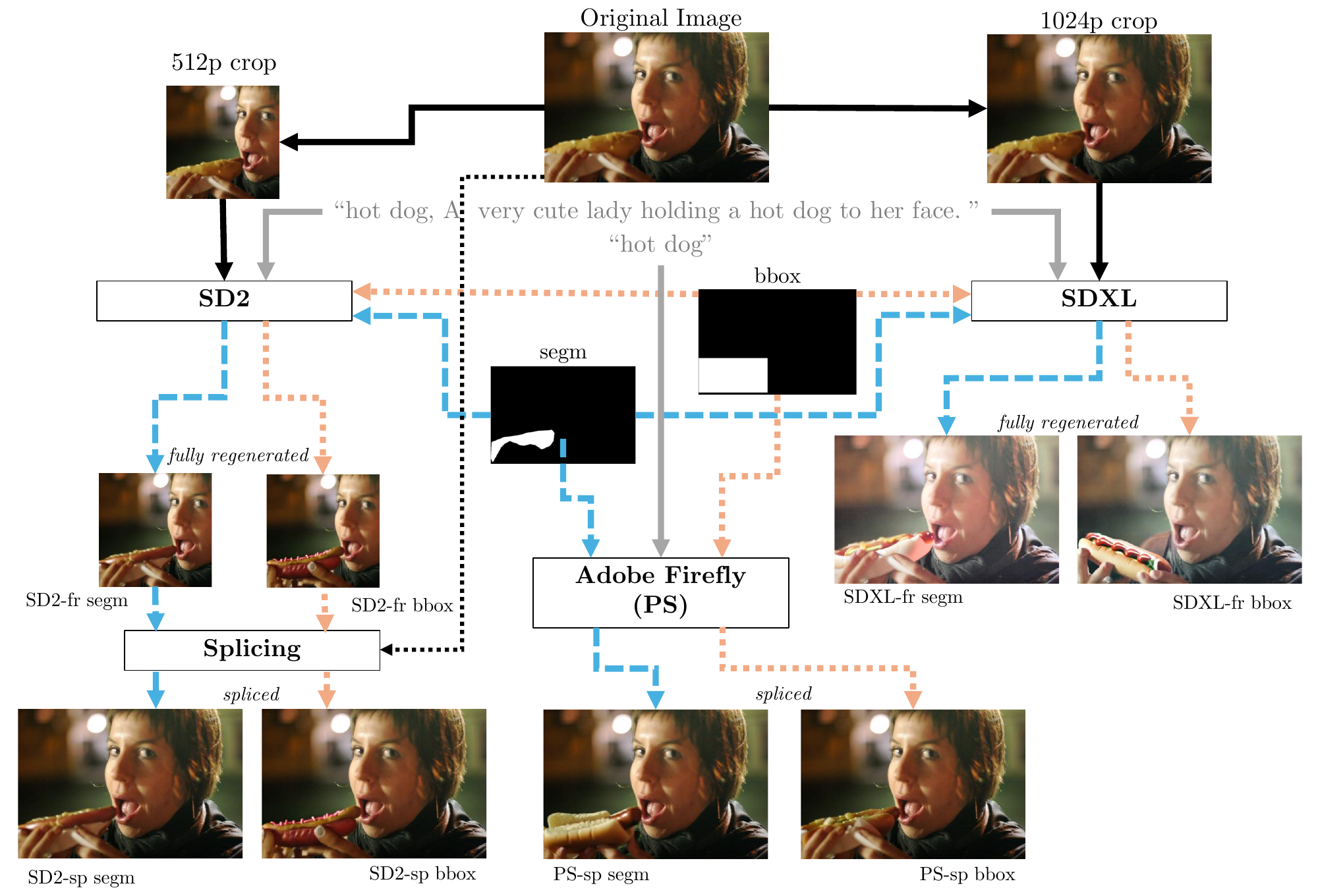}
\vspace{-0.35cm}
\caption{TGIF dataset creation. The original image is fed to SD2, SDXL, and Adobe Firefly (Photoshop), along with a prompt and a mask (segmentation or bounding box). For SD2, an additional splicing step is performed, which is done automatically in Adobe Firefly, and discarded for SDXL's output. For simplicity, only one variation is illustrated, rather than all three variations that we generated in batch.
\label{fig:dataset-diagram}}
\vspace{-0.4cm}
\end{figure*}

In summary, an authentic image is transformed into $2 \times 3 \times 4 = 24$ manipulated variants. That is, \emph{two} masks are used for its generation (segmentation and bounding box), and each generation creates \emph{three} variations in batch (\emph{num\_images\_per\_prompt}).
Using the SD2, SDXL, and PS methods, we saved either the spliced or fully regenerated image, or both. Namely, for SD2, we saved both the spliced and fully regenerated (synthetic) image crop. For SDXL, we saved only the fully regenerated (synthetic) image. For Adobe Firefly (PS), we saved only the spliced image.
As a result, we get \emph{four} sub-datasets SD2-sp, PS-sp, SD2-fr, and SDXL-fr (\emph{sp} for splicing and \emph{fr} for fully regenerated).
Fig.~\ref{fig:dataset-diagram} shows a high-level diagram of this procedure.

The dataset and code can be downloaded at \hbox{\url{https://github.com/IDLabMedia/tgif-dataset}}.

\subsection{Dataset Splits}
\label{sec:dataset-train-val-test}
We have separated the full dataset in a training, validation, and test set, in an approximate 80/10/10 setup, based on the authentic images. These sets contain $58,560$, $8,184$, and $8,232$ manipulated images ($74,976$ total), originating from $2,440$, $341$, and $343$ unique authentic images ($3,124$ total), respectively. As described in Section~\ref{sec:dataset-generation-inpainting}, each authentic image is transformed into $2 \times 3 \times 4 = 24$ manipulated images.

We have taken precaution to avoid leakage.
First, we split the dataset based on the object category. That is, the same category does not appear in multiple subsets.
Second, we deleted images that would appear in multiple subsets. That is because an image can contain multiple objects, and hence appear multiple times in the dataset (under different categories, with different masks and prompts).

\subsection{Aesthetic Quality and Image-Text-Matching Scores}
\label{sec:dataset-analysis}
Due to the automatic generation of a large dataset, the quality of the inpainted objects varies significantly. Additionally, the prompt is not always respected by the diffusion model.
To objectively measure this, we calculated aesthetic quality scores using the neural image assessment (NIMA)~\cite{talebi2018nima} (\emph{epoch-82} \& \emph{vgg16-397923af}) and generated image quality assessment (GIQA)~\cite{gu2020giqa} (\emph{pca95-cat} \& \emph{gmm-cat-pca95-full7}) metrics.
Additionally, we calculated image-text-matching (ITM) scores~\cite{li2022blip} (\emph{Salesforce\_blip-itm-base-coco}) to measure whether the prompt matches the outcome.
We measured these scores on only the inpainted area (using the provided bounding box), as opposed to on the entire image. We include these values as metadata in the dataset. 

\section{Benchmark: IFL \& SID}
\label{sec:benchmark}

\subsection{Experimental Setup}
\label{sec:benchmark-setup}
We performed the evaluation on the test set, separating the spliced subsets (SD2-sp and PS-sp) and fully regenerated subsets (SD2-fr and SDXL-fr).
Additionally, to make the task more challenging, we evaluated the performance on compressed variants of the real and fake images. That is, we used the JPEG and WEBP compression standards, using quality factors of 80 and 60.

For IFL, we use the mean pixel-level F1 score of the fake images as the performance metric, which is the harmonic mean of the precision and recall.
We utilize a threshold of 0.5 to decide whether pixels are real or fake. This threshold assesses the calibration of the models, which is essential when deploying the models in the wild and requiring human interpretation.
We selected several recent high-performing AI-based IFL methods, namely the PSCC-Net~\cite{liu2022pscc}, SPAN~\cite{hu2020span}, ImageForensicsOSN~\cite{wu2022ImageForensicsOSN}, MVSS-Net++~\cite{dong2022mvss}, Mantranet~\cite{wu2019mantranet}, CAT-Net~(v2)~\cite{kwon2022catnetv2}, TruFor~\cite{guillaro2023trufor}, and MMFusion~\cite{triaridis2024mmfusion}.

For SID,  we use the AUC score as the performance metric. 
We opted to report the threshold-agnostic AUC rather than threshold-specific image-level F1 score, because several SID methods reported high F1 scores while actually demonstrating bad performance. 
More specifically, these methods often classified both real and fake images as fake, which, due to the class imbalance, significantly boosted the F1 score.
Since the AUC represents the trade-off between the false positive and false negative rate for a range of thresholds, it better reflects detection performance, in this case.
We selected several AI-based SID methods available in the SIDBench framework~\cite{schinas2024sidbench}, namely CNNDetect~\cite{wang2019cnngenerated}, DIMD~\cite{corvi2023dimd}, Dire~\cite{wang2023dire}, FreqDetect~\cite{frank2020fredetect}, UnivFD~\cite{ojha2023univFD}, Fusing~\cite{ju2022fusing}, GramNet~\cite{liu2020gramnet}, LGrad~\cite{tan2023LGrad}, NPR~\cite{tan2024NPR}, RINE~\cite{koutlis2024Rine}, DeFake~\cite{sha2023defake}, and PatchCraft~\cite{zhong2023patchcraft}.
Some methods provide multiple models, trained on different data. We only show the best performing in this paper.
Although some of these methods were trained on GANs, some of them have demonstrated generalization performance to DMs. Hence, also methods trained on GANs were included in our evaluation set.

\subsection{Image Forgery Localization Methods}
\label{sec:benchmark-forgery}
Table~\ref{tab:benchmark-forgery} reports the F1 values for all evaluated IFL methods. We highlight F1 values above 0.7. This threshold was chosen arbitrarily, but allows to quickly grasp which methods demonstrate good detection performance.
For the splicing datasets SD2 and PS, we observe that only CAT-Net, TruFor, and MMFusion report high F1 scores.
Additionally, Mantranet only reports a mediocre F1 value for the PS dataset, while it has low performance for SD2.
All other methods demonstrate low detection performance.

For the fully regenerated datasets SD2-fr and SDXL-fr, no method is able to detect and localize these manipulations.
This is expected, since regenerating the image removes most invisible traces such as compression artifacts and camera noise.
In fact, diffusing and regenerating an image is an existing attack against forgery detection methods~\cite{Tailanian_2024_WACV}.
This highlights the threat of local image manipulations using diffusion models that fully regenerate images; these cannot be detected by traditional image forgery localization methods.
This threat is enlarged when one considers the widespread use of generative models in applications such as super-resolution~\cite{li2022srdiff} and image compression~\cite{yang2024compression}.
Although these applications can be used with harmless intentions, they may remove traces used by traditional IFL methods.

We only analyzed the influence of compression for the best performing methods in the uncompressed case, and only for the splicing datasets.
When performing JPEG compression, F1 scores drop significantly. Moreover, this performance drop is amplified when performing WEBP compression. 
This demonstrates a need to develop IFL methods that are robust to compression.

\begin{table*}[!t]
\begin{center}
\caption{Evaluation of Image Forgery Localization Methods on our TGIF Dataset (F1 Score). We only evaluated compressed images on IFL methods and sub-datasets exhibiting decent performance on uncompressed images.
\vspace{-0.25cm}
}
\label{tab:benchmark-forgery}
{
\footnotesize
\setlength\tabcolsep{4pt}%
\begin{tabular}{l cccc cc cc cc cc}
\toprule
\multirow{2}{*}{Method}
 & \multicolumn{4}{c}{Uncompressed} & \multicolumn{2}{c}{JPEG Q80} & \multicolumn{2}{c}{JPEG Q60} & \multicolumn{2}{c}{WEBP Q80} & \multicolumn{2}{c}{WEBP Q60}\\
 & SD2-sp & PS-sp & SD2-fr & SDXL-fr & SD2-sp & PS-sp & SD2-sp & PS-sp & SD2-sp & PS-sp & SD2-sp & PS-sp \\
\midrule
PSCC-Net~\cite{liu2022pscc} & 0.15 & 0.38 & 0.05 & 0.05 & - & - & - & - & - & - & - & -  \\
SPAN~\cite{hu2020span} & 0.00 & 0.00 & 0.00 & 0.00 & - & - & - & - & - & - & - & - \\
ImageForensicsOSN~\cite{wu2022ImageForensicsOSN} & 0.23 & 0.36 & 0.20 & 0.18 & - & - & - & - & - & - & - & -  \\
MVSS-Net++~\cite{dong2022mvss} & 0.07 & 0.08 & 0.06 & 0.09 & - & - & - & - & - & - & - & - \\
Mantranet~\cite{wu2019mantranet} & 0.15 & 0.56 & 0.03 & 0.05 & 0.08 & 0.13 & 0.03 & 0.04 & 0.08 & 0.11 & 0.09 & 0.11  \\
CAT-Net~\cite{kwon2022catnetv2} & \textbf{0.87} & \textbf{0.85} & 0.04 & 0.03 & 0.05 & 0.07 & 0.09 & 0.08 & 0.13 & 0.13 & 0.07 & 0.07 \\
TruFor~\cite{guillaro2023trufor} & \textbf{0.83} & \textbf{0.79} & 0.19 & 0.18 & 0.32 & 0.44 & 0.25 & 0.32 & 0.24 & 0.29 & 0.20 & 0.23\\
MMFusion~\cite{triaridis2024mmfusion} & \textbf{0.75} & \textbf{0.74} & 0.15 & 0.18 & 0.35 & 0.49 & 0.22 & 0.32 & 0.22 & 0.33 & 0.19 & 0.28 \\
\bottomrule
\end{tabular}
\vspace{-0.5cm}
}
\end{center}
\end{table*}

\subsection{Synthetic Image Detection Methods}
\label{sec:benchmark-synthetic}

\begin{table*}[!t]
\begin{center}
\caption{Evaluation of Synthetic Image Detection Methods on our TGIF Dataset (AUC Score).
\vspace{-0.25cm}
}
\label{tab:benchmark-synthetic}
{
\footnotesize
\setlength\tabcolsep{4pt}%
\begin{tabular}{ll cc cc cc cc cc}
\toprule
\multirow{2}{*}{Method} & \multirow{2}{*}{Trained on} & \multicolumn{2}{c}{Uncompressed} & \multicolumn{2}{c}{JPEG Q80} & \multicolumn{2}{c}{JPEG Q60} & \multicolumn{2}{c}{WEBP Q80} & \multicolumn{2}{c}{WEBP Q60}\\
& & SD2-fr & SDXL-fr & SD2-fr & SDXL-fr & SD2-fr & SDXL-fr & SD2-fr & SDXL-fr & SD2-fr & SDXL-fr \\
\midrule
CNNDetect~\cite{wang2019cnngenerated} & ProGAN & 0.57 & 0.61 & 0.57 & 0.62 & 0.56 & 0.61 & 0.55 & 0.60 & 0.55 & 0.60\\
DIMD~\cite{corvi2023dimd} & LDM & \textbf{1.00} & \textbf{0.94} & \textbf{1.00} & \textbf{0.95} & \textbf{0.99} & \textbf{0.92} & \textbf{0.96} & 0.71 & \textbf{0.90} & 0.61\\
Dire~\cite{wang2023dire} & ADM & 0.49 & 0.62 & 0.49 & 0.61 & 0.49 & 0.60 & 0.50 & 0.60 & 0.48 & 0.60\\
FreqDetect~\cite{frank2020fredetect} & GANs & 0.50 & 0.40 & 0.72 & 0.71 & 0.71 & 0.72 & 0.66 & 0.61 & 0.62 & 0.62\\
UnivFD~\cite{ojha2023univFD} & ProGAN & \textbf{0.82} & \textbf{0.8} & 0.73 & 0.78 & 0.70 & 0.73 & 0.71 & 0.74 & 0.69 & 0.73\\
Fusing~\cite{ju2022fusing} & ProGAN & 0.55 & 0.47 & 0.52 & 0.59 & 0.53 & 0.61 & 0.50 & 0.53 & 0.50 & 0.54\\
GramNet~\cite{liu2020gramnet} & GANs & \textbf{0.82} & 0.55 & 0.48 & 0.56 & 0.47 & 0.54 & 0.49 & 0.54 & 0.50 & 0.53\\
LGrad~\cite{tan2023LGrad} & ProGAN & \textbf{0.85} & \textbf{0.83} & 0.51 & 0.57 & 0.51 & 0.64 & 0.47 & 0.42 & 0.47 & 0.41\\
NPR~\cite{tan2024NPR} & ProGAN & 0.51 & 0.67 & 0.50 & 0.30 & 0.51 & 0.68 & 0.49 & 0.69 & 0.49 & 0.31\\
RINE~\cite{koutlis2024Rine} & ProGAN & \textbf{0.89} & \textbf{0.89} & 0.79 & \textbf{0.88} & 0.75 & \textbf{0.85} & 0.77 & \textbf{0.85} & 0.76 & \textbf{0.87}\\
 & LDM & \textbf{0.97} & \textbf{0.93} & \textbf{0.84} & \textbf{0.89} & \textbf{0.80} & \textbf{0.82} & 0.75 & 0.76 & 0.72 & 0.77\\
DeFake~\cite{sha2023defake} & Diffusion & 0.65 & 0.61 & 0.54 & 0.45 & 0.54 & 0.43 & 0.54 & 0.45 & 0.53 & 0.49\\
PatchCraft~\cite{zhong2023patchcraft} & ProGAN & \textbf{0.98} & \textbf{0.95} & 0.66 & 0.67 & 0.60 & 0.58 & 0.60 & 0.57 & 0.59 & 0.61 \\
\bottomrule
\end{tabular}
\vspace{-0.5cm}
}
\end{center}
\end{table*}

Table~\ref{tab:benchmark-synthetic} shows the performance (AUC) results for all evaluated SID methods. We highlight AUC values above 0.8. This threshold was chosen arbitrarily, but allows to quickly grasp which methods demonstrate decent detection performance.
Note that we do not include the performance results for the spliced datasets (SD2-sp and PS-sp), since we cannot expect SID to detect these types of local manipulation, as they were not designed for this purpose. We have verified that, indeed, no SID method is able to detect the synthetic spliced regions.

We observe that 6 out of 12 methods are able to detect (fully regenerated) synthetic images in uncompressed form.
Surprisingly, some methods trained on GANs (such as UnivFD, LGrad and RINE) demonstrate better performance than methods trained on DMs (such as DeFake).
Additionally, only two of those methods survive JPEG compression, and only one survives WEBP compression.
Moreover, some methods (such as DIMD) are more robust in detecting SD2 than SDXL, whereas others (such as RINE) are more robust in detecting SDXL than SD2. No single method is robust for both SD2 and SDXL on all evaluated compression levels.
This demonstrates that these methods are complementary.

In summary, we demonstrated that several state-of-the-art SID methods are able to detect fully regenerated inpainted images by SD2 and SDXL.
However, they are not able to localize the inpainted area, but rather produce a SID score for the whole image.
This is expected as they were not designed to detect the spliced inpainted images (even though the spliced regions were generated synthetically).
Finally, their performance decreases when exposed to compression, with a notable difference between WEBP and JPEG compression.

\section{Discussion \& Conclusion}
\label{sec:conclusion}
The proposed TGIF dataset represents a significant advancement in the resources available for training and evaluating forensic methods. By including high-resolution images and utilizing multiple inpainting methods, as well as providing both spliced and fully regenerated versions, our TGIF dataset addresses the limitations of existing forensic methods and datasets. As such, it is expected to contribute to the development of more effective forensic tools.

Our benchmark analysis shows that some of the existing IFL methods are able to detect and localize spliced images, whereas they fail to localize the inpainted area in fully regenerated images.
In contrast, some of the existing SID methods are able to detect fully regenerated images, yet lack the ability to localize the synthetic inpainted area.
These limitations highlight the need for new forensic methods, leveraging elements from both IFL and SID methods.
Furthermore, we showed that forensic methods suffer from a notable decrease in performance when exposed to compression, particularly with WEBP.
This highlights the need to include compressed images during training. 

Future work could extend and incorporate the TGIF dataset for the training of new detection methods to localize manipulations in fully regenerated images, with attention to robustness against compression.

%
\bibliographystyle{IEEEtran}
\bibliography{IEEEabrv,refs}

\end{document}